\documentclass[11pt, letterpaper]{article}
\usepackage[utf8]{inputenc} %
\usepackage[T1]{fontenc}    %
\usepackage[margin=1in]{geometry}

\usepackage{color}
\usepackage{url}
\usepackage{amsmath}
\usepackage[noend,boxruled]{algorithm2e}
\usepackage{caption}
\usepackage{subcaption}
\usepackage{array,multirow,graphicx}
\usepackage{bigdelim}
\usepackage{hyperref}

\DeclareMathOperator*{\argmax}{arg\,max}
\DeclareMathOperator*{\argmin}{arg\,min}

\newcommand{\para}[1]{{\vspace{2pt} \noindent \textbf{#1}
    \hspace{6pt}}}
\newcommand{\etal}{{\em et al.\ }}
\newcommand{\eg}{{\em e.g.\ }}
\newcommand{\ie}{{\em i.e.\ }}
\newcommand{\etc}{{\em etc.}}

\newcommand{\allu}{\mathcal{U}}
\newcommand{\alli}{\mathcal{I}}
\newcommand{\alls}{\mathcal{S}}
\newcommand{\iu}{\mathcal{I}_u^+}
\newcommand{\expl}{E_{i, u}}
\newcommand{\iexp}{\mathcal{I}_u^{-cf}}
\newcommand{\cfm}{\theta^{cf}}

\newcommand{\cfsym}{\mathcal{CF}}
\newcommand{\acfsym}{\mathcal{CF^A}}

\newcommand{\cffull}{{\textit{Counterfactual Proximity}}}
\newcommand{\acffull}{{\textit{Approximate Counterfactual Proximity}}}
\newcommand{\itemsim}{\textit{Item-Sim}}
\newcommand{\genresim}{\textit{Genre-Jacc}}

\title{Counterfactually Evaluating Explanations in Recommender Systems}

\author{Yuanshun Yao, Chong Wang, Hang Li\\
ByteDance\\
\texttt{\{kevin.yao,chong.wang,lihang.lh\}@bytedance.com}
}
\date{}

\begin{document}

\maketitle

\begin{abstract}
  Modern recommender systems face an increasing need to explain their recommendations. Despite considerable progress in this area, evaluating the quality of explanations remains a significant challenge for researchers and practitioners. Prior work mainly conducts human studies to evaluate explanation quality, which is usually expensive, time-consuming, and prone to human bias. In this paper, we propose an offline evaluation method that can be computed without human involvement. To evaluate an explanation, our method quantifies its \textit{counterfactual} impact on the recommendation. To validate the effectiveness of our method, we carry out an online user study. We show that, compared to conventional methods, our method can produce evaluation scores more correlated with the real human judgments, and therefore can serve as a better proxy for human evaluation. In addition, we show that explanations with high evaluation scores are considered better by humans. Our findings highlight the promising direction of using the counterfactual approach as one possible way to evaluate recommendation explanations. 
\end{abstract}

\section{Introduction}
\label{sec:intro}
Today we expect more from recommender systems than merely making recommendations. As a system that serves humans, a modern recommender system needs to meet a range of advanced responsibility requirements, including fairness, diversity, explainability, accountability, and safety \etc{} The necessity of building responsible recommender systems comes from not only the users' natural requests which lead to product demands, but also an increasing amount of moral pressure from the public opinions~\cite{youtubefairness,facebookfairness,twitterfairness} as well as compliance requirements from the government guidelines and regulations~\cite{darpa,aiuk,chinaprop}. In short, for today's practitioners,
building responsible recommender systems is commercially, morally, and legally desirable.
% building responsible recommender systems is technologically, morally, and legally required.

Driven by the needs, researchers have made considerable progress in constructing responsible recommender systems. In this paper, we focus on explainability. This is an important part of a responsible system because it offers several benefits to both users and systems. \textit{First}, it gives users more control over the system. If users know why a recommendation is made, they will learn how to change their behaviors to receive different recommendations in the future. \textit{Second}, the system can gain users' trust by showing that the system is transparent. \textit{Third}, explainability can also help practitioners to debug the recommendation model.

%Researchers have made considerable progress in building explainable recommender systems. 
For explainability of recommendation, a variety of approaches have been proposed, and can be roughly categorized as \textit{collaborative filtering based approach}~\cite{resnick1994grouplens,herlocker2000explaining,sarwar2001item} (explain by relevant users or items), \textit{user opinion based approach}~\cite{zhang2014explicit,wu2015flame,costa2018automatic} (explain by user reviews), and \textit{social interaction based approach}~\cite{sharma2013social,chaney2015probabilistic,wang2014also} (explain by user social information) \etc{} Another way to classify existing approaches is by the type of models used to generate explanations. For example, some work chooses factorization models~\cite{zhang2014explicit,zhang2015daily,chen2016learning} or neural models~\cite{seo2017interpretable,wu2019context,gao2019explainable} while others select graph models~\cite{heckel2017scalable,he2015trirank,wang2018tem} or natural language models~\cite{chang2016crowd,wang2018explainable,costa2018automatic}.

Prior work so far focuses on developing algorithms that generate explanations, while little attention is paid to evaluating the quality of generated explanations. Because eventually recommendation explanations are judged by users, prior work relies largely on conducting human studies to assess how well their proposed algorithms perform. However, for a recommender system deployed in the real world, conducting user studies is a costly and painful process. \textit{First}, it is slow to collect users' feedback and therefore would become the bottleneck of the algorithm development. \textit{Second}, it is expensive since online human assessment costs more than offline algorithm testing. \textit{Third}, it is prone to numerous human biases, \eg methodological bias, familiarity bias, and culture bias \etc{} In summary, suitable evaluations are vital to the development of explainability in recommender systems, and it is currently missing in this research area.

To overcome the challenge, we propose an offline evaluation method that can be computed without human involvement. It is fast, low-cost, scalable, unaffected by human bias, and therefore more suitable for evaluating explanations in real-world recommender systems. We design the evaluation method based on \textit{counterfactual} logic, following the insights from \textit{counterfactual psychology}~\cite{kahneman1981simulation, mandel2005psychology,carpenter1973extracting,fillenbaum1974information} and \textit{counterfactual machine learning}~\cite{wachter2017counterfactual,mothilal2020explaining,russell2019efficient}. Our method produces a score that measures the quality of an explanation by quantifying the explanation's \textit{counterfactual impact} on the recommendation. Through an intensive user study ($N=400$), we show that the evaluation scores are closely correlated with the human judgments, and therefore can serve as a good proxy for the evaluation of explanations. In addition, we demonstrate that the explanations with high scores are generally considered better by humans than those from the baseline approaches.
We summarize our contributions as follows:
\begin{itemize}
   \item We propose a new method that evaluates the quality of explanations in recommender systems using counterfactual logic.
    % \item Through an extensive user study, we validate the effectiveness of  our evaluation method.
    \item Through an extensive user study, we validate the effectiveness of the evaluative scores produced by our method by demonstrating its close correlation to human judgments.
    \item We show that humans consider the explanations with high evaluation scores better.
\end{itemize}

We hope that our work can bring more attention to research on evaluation methods for explainability in recommender systems. Our counterfactual design is one possible way to approach the problem, and there might be other ways to design evaluation metrics. Algorithm development requires a variety of evaluation metrics, and the more the better. As we see from other areas, \eg \textit{BLEU score}~\cite{papineni2002bleu} in \textit{natural language processing} and \textit{NDCG measure}~\cite{al2007relationship} in \textit{information retrieval}, a good evaluation method can greatly accelerate the advancement of a field.
\section{Preliminary}
\label{sec:back}
In this section, we begin by providing some background information on explainable recommender system, counterfactual machine learning, and the design choice of using counterfactual logic to assess recommendation explanations.

\subsection{Explainable Recommender System}

\para{Why Explainable Recommender System?} An explainable recommender system offers several benefits to both users and the system. \textit{First}, it provides users with more control over the system. If a user is unhappy with the current recommendation, the explanation can tell users what actions to take to change future recommendations. Tran \etal~\cite{tran2021users} show that users are more likely to want explanations when they dislike the current recommendation. In this case, by choosing to interact with items different from those in the explanation, users can stop receiving similar recommendations in the future. \textit{Second}, explainability might help the conventional goals of recommendation, \eg knowing why an item is recommended might increase the chance for users to interact with the item. \textit{Third}, the system can gain user trust by showing the system is transparent through explanations. \textit{Fourth}, explainability can also help practitioners debug the recommendation models. \textit{Finally}, it fulfills the potential legal requirements as explainability has been increasingly mentioned in governmental AI proposals from various countries including US~\cite{darpa}, UK~\cite{aiuk}, and China~\cite{chinaprop}.

\para{Current Approach.} In the following, we briefly summarize the major categories of existing approaches in explainable recommender system. For more details, refer to ~\cite{zhang2018explainable} for a general survey.
\begin{itemize}
    \item \textit{Collaborative filtering based logic}: The logic uses relevant users or items to explain recommendations. In \textit{user-based collaborative filtering}~\cite{resnick1994grouplens,herlocker2000explaining} logic, the recommender system finds a group of users similar to the current user, and explains that the system recommends the current item because some neighborhood users had some positive feedback on it. On the other hand, \textit{item-based collaborative filtering}~\cite{sarwar2001item} logic explains the recommendation by finding a group of items from the current user's interaction history and then informing the user the recommendation reason is that the user has interacted with those similar items in the past. For both logic, the relevance is usually obtained from user-item interaction. But broadly speaking, it can also be defined by features of items or users, \eg user demographic information~\cite{pazzani1999framework,zhao2014we,zhao2016exploring} or item features~\cite{vig2009tagsplanations,cramer2008effects,pazzani2007content,hou2019explainable}.
    \item \textit{User opinion based logic}: The logic explains the recommendation through user opinions, mostly from user reviews. Most of the work is based on natural language processing techniques. For example, some prior work uses topic modeling to generate word clouds as explanations ~\cite{zhang2014explicit,wu2015flame}. In addition, using natural language generation, Costa~\etal~\cite{costa2018automatic} train language models on user review corpus, and generate explanations in the form of natural language paragraphs based on the learned language models. Chang~\etal~\cite{chang2016crowd} generate personalized natural language explanations based on explanations collected from crowdsourcing. Furthermore, Wang \etal~\cite{wang2018explainable} propose a multi-task learning modeling framework that predicts how the user would appreciate a particular item during training.
    \item \textit{Social interaction based logic}: The logic generates explanations based on the users' social information. Sharma and Cosley show users a number of friends who also liked the item~\cite{sharma2013social}. Chaney~\etal~\cite{chaney2015probabilistic} incorporate social influence information from social network into the model, and provide explanations by indicating friends as the source of influence. Wang~\etal~\cite{wang2014also} find a group of users who are the most persuasive in terms of explaining, and provide explanations in the text form ``A and B also like the item.''
\end{itemize}

Note that different explaining logic would require different explaining user interfaces. For example, \textit{collaborative filtering} logic displays the relevant users or items, \textit{user opinion} logic shows a text block, and \textit{social interaction} logic has different user interface designs, ranging from a text sentence/paragraph to links to source influence. Because of their difference in displaying forms, it is difficult to compare explanations generated from different explaining logic in a controlled experimental setting. In this work, we focus on \textit{item-based collaborative filtering} logic.

\para{Evaluate Explanation Performance.} To the best of our knowledge, most of the prior work evaluates their explanation quality using online user study and only a few works consider offline evaluation. For example, Abdollahi and Nasraoui~\cite{abdollahi2017using} propose evaluation metrics based on the expected value of ratings given by similar users. The majority of prior work relies on conducting human studies, which is slow to collect results and expensive and therefore not suitable for rapid algorithm development and testing needed in today's recommender systems. In addition, user studies are also prone to various human biases, \eg methodological bias, familiarity bias, and culture bias \etc{} Since reliable and low-cost evaluation metrics can be beneficial by helping practitioners develop and test explaining algorithms, we try to fill this gap in this paper.

% \vspace{-0.1cm}
\subsection{Counterfactual Machine Learning}
The term \textit{counterfactual} refers to a way of thinking which assumes scenarios that contradict the previous facts, \ie ``What-if'' scenarios. Recently machine learning researchers have been studying how to use counterfactual logic to explain model predictions. A typical example is a loan application. Banks develop models to predict if a user's loan application is approved or rejected based on the user's features like income, age, credit card history \etc{} If a user's application is rejected by the model, a counterfactual explanation would be given to the user by pointing out what changes on which features would lead to an approval, \eg ``Had you earned $\$10,000$ dollars more annually, we would approve your application.'' 

Compared to other explainable machine learning approaches, counterfactual explanation has several advantages. \textit{First}, it provides users actionable advice, \ie users would know how to change the model decisions, and can try the advice if needed. \textit{Second}, it might be more likely to be accepted by humans because we naturally think counterfactually in everyday life~\cite{kahneman1981simulation, mandel2005psychology}. \textit{Third}, counterfactual explanations are technically verifiable in truthfulness. If we make a counterfactual claim, we can test if this is true or not in our models. On the other hand, other types of explanations are hard to validate the truthfulness of the claimed reasons in reality. 

\para{Current Approach.} Wachter~\etal~\cite{wachter2017counterfactual} propose one of the first widely used approaches to counterfactually explain model predictions. The idea is to use white-box optimization on models to find the smallest change in an input's features that can alter the model prediction on this input. Mothilal~\etal~\cite{mothilal2020explaining} consider the need to generate multiple explanations, and argue they should be different from each other. To this end, they add a diversity loss into the optimization that maximizes the diversity between explanations. Russell~\etal~\cite{russell2019efficient} propose a mixed integer programming based approach to address the difficulty of generating sensible explanations regarding categorical features. In addition to generic models, counterfactual explanations are also designed in applications like Computer Vision~\cite{dhurandhar2018explanations,goyal2019counterfactual,akula2020cocox}, Natural Language Processing~\cite{wu2021polyjuice,yang2020generating}, and Graph Neural Networks~\cite{ying2019gnnexplainer,yeh2019diverse}. See~\cite{verma2020counterfactual} for a literature review.

Relatively fewer works consider applying the counterfactual approach to generate recommendation explanations. Our work is close to this line of research although our goal is to evaluate explanations rather than generate explanations. Ghazimatin~\etal~\cite{ghazimatin2020prince} consider graph recommender models that encode user-item interactions. The counterfactual explanation is defined as a set of minimal actions performed by the user that, if were removed, changes the recommendation to a different item. They propose a Personal PageRank based approach that searches for the minimal change on the graph. However, this approach is only applicable to PageRank-based recommender models, and cannot be easily adapted to other model categories. In addition, Tran~\etal~\cite{tran2021counterfactual} propose a white-box explaining approach on Neural-CF models. 
% The definition of counterfactual explanation is a set of items that, if were removed from the user's interaction history, would change the current recommendation. 
The approach is based on the influence function~\cite{koh2017understanding} which estimates the influence of a training sample on the current model prediction. They use the influence function to compute how the predicted score on the explained item might change if some training items were removed.
% Then they greedily add training items to the solution set in the order of decreasing change in predicted score. 
Furthermore, Kaffes~\etal~\cite{kaffes2021model} propose a black-box solution that performs Breadth First Search with heuristics that combine search length and drop of the item rank if the candidate set were considered.

% \kevin{mention pearl's work}
\subsection{Why Use Counterfactual Logic to Evaluate Explanations?}
\label{sec:why}
Counterfactual thinking involves ``What-if'' reasoning that is contrary to prior facts, \eg ``What if I did that instead?''. Under the explainable recommender system setting, one possible counterfactual explanation can be ``Had you not interacted with those items, you would not receive this recommendation.'' Psychological research has found counterfactual thinking is a frequent pattern of human reasoning in everyday activities~\cite{kahneman1981simulation, mandel2005psychology}. Therefore not only it provides a possible way to model human reasoning, but also this frequently used thinking pattern is already familiar to humans and therefore is likely to be understood and accepted by users. It naturally represents a common pattern of human reasoning in everyday life, and can serve as a proxy for human judgments. If an explanation is close to being \textit{counterfactual}, then it is more likely to fit the pattern of human counterfactual thinking, and make users satisfied.

\section{Methodology}
\label{sec:method}
In this section, we first introduce our problem scope and setup, followed by the demonstration of our key design insights. We then present the details of our explanation evaluation metric.

\subsection{Problem Scope}
We focus on evaluating the quality of recommendation explanations generated based on \textit{item-based collaborative filtering} logic, which is to explain the current recommendation by its relevant items, \ie the model recommends the item because the item is similar to some other items that the user has interacted with previously. The explanation user interface shows a set of similar items in the user's history with an explanation text like ``You have interacted with those similar items before.'' or ``You have rated high on those similar items before.'' We focus on this explanation logic for two reasons. \textit{First}, it is simple and therefore easy to implement within the existing recommender systems. It does not require extra models and can be implemented with a small engineering module. \textit{Second}, it can be applied to any recommendation model because we can always find relevant items based on criterion or distance metrics defined independently of the recommendation model (\eg item raw features). Note that the explanations generated based on other logic, \eg \textit{user-based collaborative filtering}, \textit{user opinion}, or \textit{social interaction}, are not comparable in our setting because they would require different algorithm design and explaining user interface, and therefore cannot be applied to our scenario under a controlled experimental setting.

\subsection{Problem Setup}
In a recommender system, let $\allu = \{u_1, \dots, u_m\}$ be the set of all users, $\alli = \{i_1, \dots, i_n\}$ be the set of all items, and $\alls \subseteq \allu \times \alli$ be the history of all user-item interactions. For a given user $u$ who has interacted with the item set $\iu =\{i \in \alli: (u, i) \in \alls\}$ in the past, if the recommendation model $\theta$ recommends an item $i$ to the user $u$, then the goal of the explaining component of the system is to generate explanations of why the item $i$ was recommended to the user $u$. This is done, in an \textit{item-based collaborative filter} logic, by showing user $u$ a subset of his or her interaction history $\expl \subseteq \iu$ as the explanation.

Given an explanation $\expl$, the goal of an offline explanation evaluation task that we target is to evaluate $\expl$'s quality by computing a numerical score on $\expl$ which is correlated with human judgments if a human were to see $\expl$.

\begin{figure}[!t]
  \centering
  \includegraphics[width=0.8\linewidth]{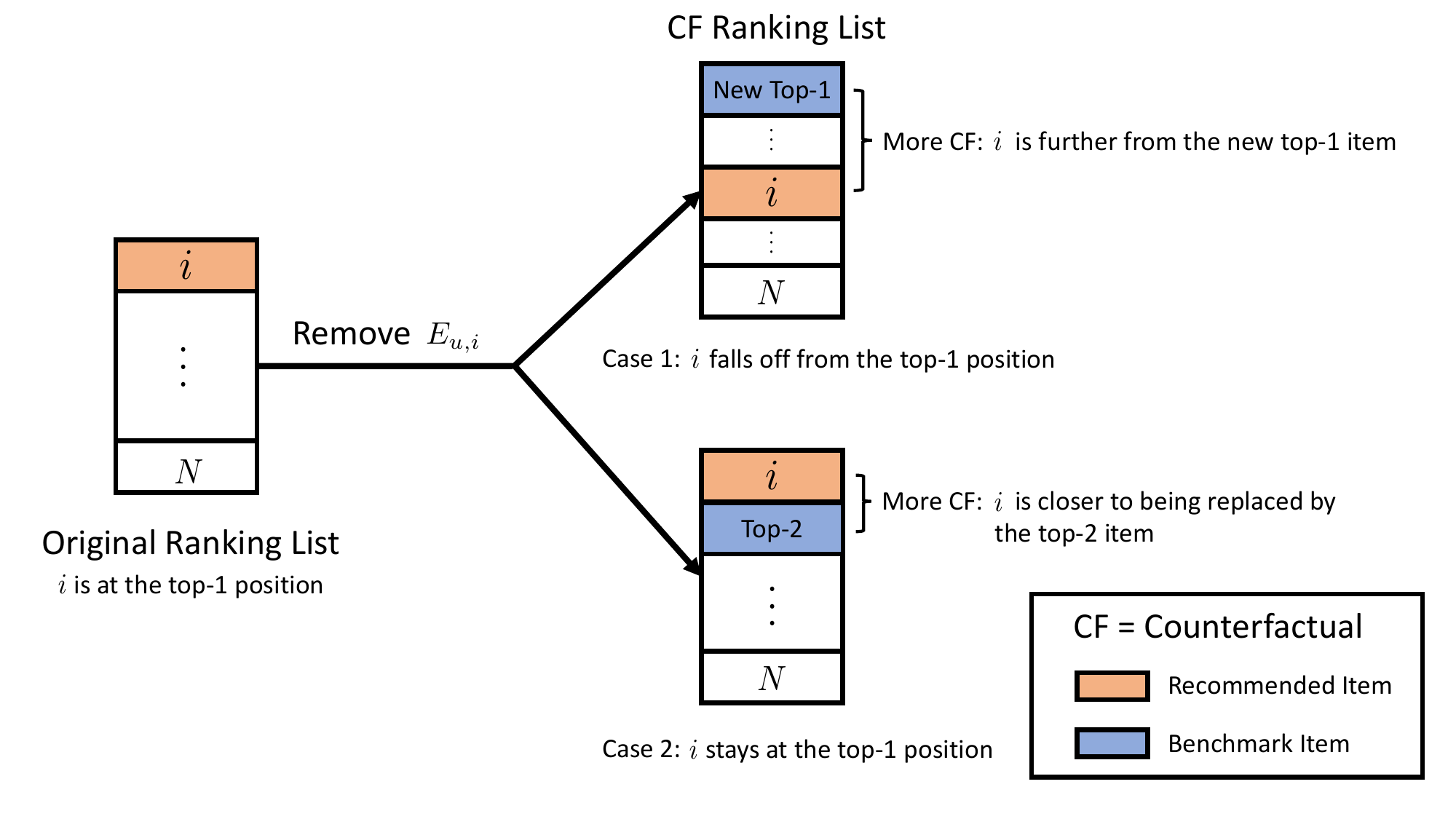}
  \caption{The key insight on how to compute \cffull. Suppose item $i$ is originally ranked at the top-1 position. Then we remove $\expl$ from the model's training data and retrain the model to obtain a new ranking list, \ie the counterfactual ranking list. In the counterfactual ranking list, the less likely for the recommended item $i$ to remain at the top-1 position, the larger impact removing $\expl$ would have on the recommendation of item $i$. In other words, item $i$ would be more likely to disappear from the recommendation if the user had not visited $\expl$ before. Therefore we should assign a higher counterfactual score to $\expl$ if after removing it, item $i$ is more likely to be replaced.}
  \label{fig:cf_insight}
\end{figure}

\subsection{Design Insights}
We now describe the key design insights of our counterfactual evaluation method.

\para{Qualitative Definition.} We define an explanation $\expl$ to be counterfactual if $\expl$ were removed from user $u$'s interaction history $\iu$, and the model were trained on the removed training set $\alls \setminus \{u \times \expl\}$ instead, then the counterfactually trained model $\cfm$ would no longer recommend item $i$ to user $u$. 

Mathematically, a counterfactually trained model $\cfm$ w.r.t $\expl$ is defined as:
\begin{equation}
\cfm = \argmin_{\theta \in \Theta}{\sum_{(x, y) \in S \setminus \{u \times E_{u, i}\}}{\ell (x, y; \theta)}}
\end{equation}

where $\Theta$ is the model parameter space and $\ell(x, y; \theta)$ is the loss of model $\theta$ on the user $x$ and the item $y$.

To simplify notations, define $\mathcal{I}_u^- = \mathcal{I} \setminus \iu$, \ie items without interaction or explicit positive feedback from the user $u$ in the training data and therefore are available for the current recommendation (because usually a recommender system does not recommend items that have already been seen by the user before), and $\iexp = \mathcal{I}_u^- \cup \expl$, \ie items available to be recommended to user $u$ in the counterfactual case (because $\expl$ now would be available for recommendation after it is removed). Let $f(u, i; \theta)$ be the predicted score of item $i$ for user $u$ by the model $\theta$, then an explanation $\expl$ is qualitatively counterfactual if
\begin{equation}
\bigg [\argmax_{j \in \iexp}{f(u, j; \cfm) \bigg ] \neq i}
\label{eq:cfdef}
\end{equation}

% That is to say, if $\expl$ were removed from user $u$'s history, the model, unlike the previous recommendation, would no longer recommend item $i$ to user $u$.

\para{Quantitative Measure.} Note that definition (\ref{eq:cfdef}) is either satisfied or unsatisfied. However, in practice it is better to have a quantitative metric which is a continuous score that measures \textit{the closeness of an explanation to its counterfactual state}. In other words, we would like to have a continuous spectrum that \textit{captures the degree of how counterfactual it is}. It is because a continuous metric is easier for practitioners to compare. Considering two explanations that both satisfy the qualitative definition of being counterfactual, we would like to know which one is more counterfactual, \ie which one, if were removed, would have a larger impact on the current recommendation. A continuous degree of counterfactuality would capture if an explanation fits with the counterfactual state on a finer granularity. We name this quantitative metric \cffull{}.

Figure~\ref{fig:cf_insight} shows the key insight on how to compute \cffull. Suppose the recommended item $i$ that we want to explain is ranked at the top-1 position. Then we remove $\expl$ from the model's training data and retrain the model to generate a new item ranking list (to the user $u$), \ie the counterfactual ranking list. In the counterfactual ranking list, the less likely for the recommended item $i$ (which is originally top-1) to stay at the top-1 position, the larger impact removing $\expl$ would have on the recommendation of item $i$ (by pulling the item $i$ away from the top-1 position and therefore replacing the recommendation by a different item), and therefore we should assign a higher counterfactual score to $\expl$. 
% In other words, \textit{the further item $i$ is from the top-1 position in the counterfactual ranking list, the more counterfactual $\expl$ is}.

\subsection{Evaluation Score: Counterfactual Proximity}

With the above insight in mind, there are two different cases depending on the position of item $i$ in the counterfactual ranking list (as shown in Figure~\ref{fig:cf_insight}): either item $i$ falls off from the top-1 position, \ie $\expl$ is qualitatively counterfactual; or item $i$ stays on the top-1 position, \ie $\expl$ is not qualitatively counterfactual. We can quantify the counterfactual impact in the former case by the score difference between item $i$ and the item ranked at the current top-1 position (\ie top-1 item), and quantify the latter case by the score difference between item $i$ and the item ranked at the current top-2 position (\ie top-2 item).
% \begin{itemize}
% \item \textbf{Qualitatively Counterfactual Case}: If item $i$ falls off from the top-1 position, \ie $\expl$ is qualitatively counterfactual that is able to change the prediction if $\expl$ were removed, then we can quantify how counterfactual $\expl$ is by observing the distance between item $i$ and the item ranked at the current top-1 position (\ie top-1 item). The further item $i$ is from the top-1 position, the more counterfactual $\expl$ is. The distance can be specified by the predicted score difference between top-1 item and item $i$:
% \begin{equation}
% \bigg [\max_{j \in \iexp}{f(u, j; \cfm)} \bigg ] - f(u, i; \cfm)
% \end{equation}
% The score is positive, and the larger means the more counterfactual $\expl$ is.
% \item \textbf{Qualitatively Non-counterfactual Case}: If item $i$ stays on the top-1 position, \ie $\expl$ is not qualitatively counterfactual, then we can quantify how counterfactual $\expl$ is by observing the distance between item $i$ and the item ranked at the current top-2 position (\ie top-2 item). The closer item $i$ is from the top-2 item, the more likely item $i$ would be replaced by the top-2 item, and therefore more counterfactual item $i$ is. The distance can be specified by the predicted score difference between top-2 item and item $i$:
% \begin{equation}
% \bigg [\max_{j \in \iexp \setminus \{i\}}{f(u, j; \cfm)} \bigg ] - f(u, i; \cfm)
% \end{equation}

% The score is non-positive, and the larger (\ie closer to 0) means the more counterfactual $\expl$ is.
% \end{itemize}

We can combine both cases. Define \textit{benchmark item} to be the item other than item $i$ that is ranked highest in the counterfactual ranking list. If item $i$ is no longer top-1, then the benchmark item is the new top-1 item; otherwise, it is the top-2 item below item $i$. When $\expl$ is more counterfactual, it implies that item $i$ is closer to being ranked lower than any other items (\ie closer to being replaced by an item), which is equivalent to being ranked lower than the benchmark item. Therefore \textit{the counterfactual score can be quantified by the score difference between the benchmark item and item $i$}. To sum up, we compute \cffull{} ($\cfsym$ in short) by the following:

\begin{equation}
\begin{aligned}
& \cfsym(u, i, \expl, S) = \bigg [\max_{j \in \iexp \setminus \{i\}}{f(u, j; \cfm)} \bigg ] - f(u, i; \cfm) \\
& \text{where } \cfm = \argmin_{\theta \in \Theta}{\sum_{(x, y) \in S \setminus \{u \times E_{u, i}\}}{\ell (x, y; \theta)}} 
\end{aligned}
\label{eq:cf}
\end{equation}

\para{Properties of \cffull.} Assume each predicted score is normalized between 0 and 1, then $\cfsym \in [-1, 1]$. The sign of $\cfsym$ has the following meaning: if it is positive, then it implies that $\expl$ satisfies the qualitative counterfactual definition; otherwise, it does not satisfy. The higher $\cfsym$ is, the more counterfactual $\expl$ is, and therefore is considered to be a better explanation by the counterfactual logic.

\para{Remarks.} Note that the above definition focuses on how the top-1 recommended item would change counterfactually because we assume the recommender system only shows the top-1 recommended item to users. One can easily extend it to top-K recommended items if one is interested in the scenario in which users are shown multiple recommended items from the ranking list simultaneously. In addition, there might exist other counterfactual definitions, \eg one can look at the rank difference rather than score difference. It is impossible to explore and examine all of them in depth within the page limit of a single paper.

\subsection{Approximate Counterfactual Proximity}
\cffull{} can be expensive to compute because it requires retraining the model. In practice, it might be too costly for large models. To speed up, we can approximate it by not fully retraining the model. To this end, we can perform a warm-start from the previous model weights, and finetune the model on the counterfactual training dataset. Since the previous model has already converged, this usually requires much fewer iterations to reach a good state. After we have this approximated version of the counterfactual model, we can compute the score by the same Eq (\ref{eq:cf}). We refer to the resulting score as \textit{\acffull} or $\acfsym$ in short. It can be viewed as a faster but less accurate version of \cffull.

\para{Computational Cost.} We admit that, in some large-scale settings, even finetuning in \textit{\acffull} might be expensive to perform. This is the major limitation of our approach. As the first paper that demonstrates the possibility of using the counterfactual approach as an evaluation, our main focus is its effectiveness in serving as a proxy for human judgments (which will be shown in the next sections) rather than speed optimization.If the direction seems promising, then in the future, the community can work towards algorithm speedup. In fact, the problem of how to obtain the counterfactual models efficiently without full retraining is the exact topic of the ongoing research area of \textit{machine unlearning}~\cite{cao2015towards,bourtoule2021machine,gupta2021adaptive,guo2019certified}. For example, one simple speedup is to replace the retraining with influence function~\cite{koh2017understanding}. We believe as machine unlearning algorithms advance, our evaluation method will become more and more efficient. In addition, compared to the current approach of conducting online user studies, our method is still relatively fast.

\section{User Study: Design}
\label{sec:human}
To understand the evaluative power of our proposed score, we ask the following research questions:

\begin{itemize}
    \item \textbf{RQ1}: Is the proposed score correlated with real human judgment?
    \item \textbf{RQ2}: Can the proposed score be used to predict human judgment?
    \item \textbf{RQ3}: Is an explanation with a high proposed score considered good by humans?
\end{itemize}

To understand the above questions, we design and conduct a survey-based user study to collect user ratings on explanations that serve as the ground truth to assess our evaluation method. In this section, we first introduce our overall design, followed by how we generate the survey data, how we ask users to rate explanations, and how we collect responses.

\subsection{Overall Design}
% In this subsection, we introduce the overall design of our user study, survey questions, quality-control mechanisms, and how we generate the survey data.
We choose movie recommendation to be our user study scenario because it is one of the most widely used cases that study recommendation explanation in the literature~\cite{tran2021users,resnick1994grouplens,herlocker2000explaining}.
In our survey, we first ask users to imagine that they are browsing a movie recommendation website. Then we show them a list of questions. In each question, we display a group of movies as the hypothetical history that users imagine having watched in the past. Next, we present them with the corresponding recommended movie based on the history (generated from the recommendation model). Then we provide the recommendation explanation based on the \textit{item-based collaborative filtering} logic, \ie ``We recommend this movie to you because you watched the following movies in the past.'' And we show users a set of movies from history as the explanation. We will explain how we choose the explanation in Section~\ref{subsec:generate}. Finally, we ask the users to rate the explanation based on various assessing dimensions introduced in Section~\ref{subsec:rating}. We repeat the above process and include 12 questions totally in the survey.

\subsection{Survey Data Generation} 
\label{subsec:generate}
We now introduce how we generate data used in the survey question.

\para{Dataset and Model.} We use MovieLens (ml-latest-small)\footnote{\url{https://files.grouplens.org/datasets/movielens/ml-latest-small-README.html}} as our dataset because it is one of the most popular datasets for the movie recommendation scenario. It contains $\sim$600 users, $\sim$9.7K items, and $\sim$101K ratings. We use Matrix Factorization (MF) model as our recommendation model since it is a naturally suitable choice for \textit{collaborative filtering} based explanations~\cite{resnick1994grouplens,herlocker2000explaining}.

\para{User History Generation.} To ensure the movies are likely to be familiar to participants, we select movies in the hypothetical watching history presented to the participants from popular movies. Specifically, we first compute the distribution of each movie's popularity (number of users who watched/rated the movie) in MovieLens, then we take the $90\%$ quantile of that distribution, and use it as the minimum popularity threshold to select movies presented as the hypothetical watching history.\footnote{We admit that the popularity filtering might cause bias because the resulting history only represents the scenario in which users watched popular movies in the past. But we think the benefit that participants would be familiar with the movies and therefore provide impartial feedback outweighs the potential risk of bias. In addition, the scenario is likely to hold in practice for the majority of users. We leave considering long-tailed users as future work. In addition, this popularity filtering only applies to user study samples rather than training samples.} After the filtering, we randomly choose 9 movies as watching history in each question.\footnote{We admit that assuming a fixed-size history might cause bias in our design. However, it is difficult to include all possible scenarios that one might encounter when using a recommender system while maintaining a statistical significance given a limited amount of resources.}

\para{Recommendation Generation.} We train the MF model on the MovieLens dataset for 20 iterations with embedding size 40. Then we generate the recommended movie by feeding the hypothetical user watching history into the trained MF model, and choose the movie with the highest predicted score as the recommended movie shown in the survey question.

\para{Compare Evaluation Scores.} To measure how well our proposed evaluation scores perform when evaluating explanation quality, we include two baseline scores to compare with: item similarity and movie genre overlap. Those two scores are relatively simple because, to the best of our knowledge, few prior works have considered proposing offline evaluation scores and we cannot find more developed evaluation methods that fit into our scenario. In the following, we explain how we compute each score in the survey. Note that we assume an explanation $\expl$ is given as the input to the evaluation method, and we will explain how we choose those explanations shortly after.

\begin{itemize}
    \item \cffull{}: Given an explanation $\expl$ for user $u$ to explain the recommended item $i$, we compute the evaluation score based on Eq (\ref{eq:cf}). We follow the same configurations as the original training to retrain the model on the counterfactual dataset (\ie rating matrix after removing interactions between user $u$ and explaining item set $\expl$) and obtain the counterfactually trained model $\cfm$.
    \item \acffull{}: Instead of fully retraining the MF model, we finetune the model by \textit{Alternating Least Square} (ALS)~\cite{als}. ALS trains an MF model by alternating between 1) fixing the item factor while updating the user factor, and 2) fixing the user factor while updating the item factor. In our case, we only run the first half of ALS, \ie we fix the item factor, and only update the user factor on the counterfactual training set. (The reason is item embeddings should be relatively unaffected by removing a few interactions from one user's history while user embeddings that include the particular user $u$ should be impacted more.) % We run this update for 5 iterations.
    \item \textit{Item Similarity} (baseline): The goal is to quantify how close the explanation $\expl$ is to the recommended item $i$ in the item embedding space. To this end, we compute the cosine similarity between the item embedding of each explaining item in $\expl$ and the recommended item $i$, then we average the similarities over all items in $\expl$. The larger is better. We name this score as \itemsim{} for short.
    \item \textit{Movie Genre Overlap~\cite{vig2009tagsplanations}} (baseline): The goal is to quantify how close the explanation $\expl$ is to the recommended item $i$ in terms of movie genre's overlap. To this end, we compute the Jaccard index between the movie tags of each explaining item in $\expl$ and the recommended item $i$, then we average the Jaccard similarity over all items in $\expl$. The larger is better. We name this score as \genresim{} for short.
\end{itemize}

\para{Choose Explanations.}  We now explain how we choose the explanations shown in our survey. To answer RQ1 and RQ2 which are about the relationship between evaluation score and human rating on the explanations, we do not need to consider what kind of explanations are used. This is because what we want to assess is the quality of the score that serves as an evaluative function rather than the quality of the explanations. Hence explanations produced by any method are usable. However, to answer RQ3, we would need to have explanations with different ranges of evaluation scores so that we can compare the rating difference between explanations with different score levels. In summary, RQ1 and RQ2 are indifferent to which explanations are chosen while answering RQ3 needs to have explanations that cover a range of different score levels. Therefore we choose to use explanations with different levels of evaluation scores.

With the above insights in mind, we select explanations with three different ranges for each score. We fix the explanation size to be 3 items. For $\cfsym$ and $\acfsym$, we enumerate 3-item subsets of the history as the explanation candidates. Then we compute $\cfsym$ or $\acfsym$ on each candidate and pick three candidates with the highest, the lowest, and (closest to) the average score respectively as the final explanations. For ~\itemsim{} and ~\genresim, we compute the score on all items in the history, and select 3 items with the highest, the lowest, and (closest to) the average score respectively. In total, for each of the four evaluation scores, we have three explanations chosen at different score levels. Hence we have 12 explanations in total; one for each question. For each chosen explanation, we also compute the other three scores so that we can compare them together. 

% Note that we do not use our evaluation to evaluate the explanations generated by our evaluation score. Because again, we are interested in assessing the evaluation method rather than the explanations, and it does not matter how explanations are generated since we only need their human judgements and their evaluation scores.

% note that we do not

% we include four types of explanation generation: $\cfsym$, $\acfsym$, ~\itemsim, and ~\genresim. . For each generation method, we use the exhaustive search to find all candidates, and compute the corresponding evaluative score. For each of 4 evaluative scores, we take the highest, lowest, and mean candidate as the explanation, and in total we have 12 questions. all of them are computed with the evaluation score.

% We use the same MovieLens dataset and Matrix Factorization model as introduced in Section~\ref{sec:data} to generate survey data. 
% In each question, there are three types of data to generate: hypothetical user watching history, recommended movie, and recommendation explanation. 

\subsection{Explanation Rating Question} 
\label{subsec:rating}
We ask users to give 1-5 ratings on the explanations (without revealing how we choose them), and use them as the ground truth of the explanation quality. Although there exist some other methods to assess explanation quality, \eg online A/B test~\cite{zhang2014explicit,sayres2019using} or asking users to predict the model behaviors on the unseen instances~\cite{ribeiro2018anchors}, we choose this method because it is one of the most common explanation evaluation protocols in the literature of explainable recommender system~\cite{herlocker2000explaining,vig2009tagsplanations,ren2017social,tran2021users}. Following the prior work~\cite{tintarev2007survey,zhang2018explainable,tran2021users,wang2018explainable}, we design questions from the following seven  assessing dimensions:
\begin{enumerate}
    \item \textbf{Explainability}: Given a scale from 1 to 5, does the explanation help you understand why we recommend this movie to you? (1 is the least helpful, 5 is the most helpful)
    \item \textbf{Informativeness}: Given a scale from 1 to 5, does the explanation help you know more about the recommended movie? (1 is the least helpful, 5 is the most helpful)
    \item \textbf{Effectiveness}: Given a scale from 1 to 5, does the explanation help you decide whether you want to watch this recommended movie? (1 is the least helpful, 5 is the most helpful)
   \item \textbf{Persuasiveness}: Given a scale from 1 to 5, does the explanation convince you to watch the recommended movie? (1 is the least convincing, 5 is the most convincing)
   \item \textbf{Transparency}: Given a scale from 1 to 5, does the explanation help you understand how the system works? (1 is the least helpful, 5 is the most helpful)
   \item \textbf{Trustworthiness}: Given a scale from 1 to 5, does the explanation increase your trust in the system? (1 is increasing the least, 5 is increasing the most)
   \item \textbf{Satisfaction}: Given a scale from 1 to 5 does the explanation make you happier about the system? (1 is the least satisfactory, 5 is the most satisfactory)
\end{enumerate}

In addition to those rating questions, we also ask users, in some of the questions, to write down the reason why they give such ratings so that we can analyze their reasoning in our study.

\subsection{Response Collection}
% We summarize some statistics of our survey response collection, followed by analyzing effectiveness of our methods.
\para{Quality Control.} To ensure response quality, we use five mechanisms to filter out low-quality responses. \textit{First}, we select workers with at least an $80\%$ approval rate in participating history (HIT rate on Amazon MTurk) and with at least $50$ tasks approved in the past. \textit{Second}, to make sure participants are reasonably familiar with movies, we declare, in the advertisements, the requirement of movie familiarity for the participants. \textit{Third}, in the middle of our survey, we insert a quality control question (``Please choose both A and D.''). Participants who fail to respond properly to this simple question are excluded from the results. \textit{Fourth}, at the end of the survey, we ask participants how many movies they usually watch in a month, and we remove any participants who answer less than one movie per month. \textit{Finally}, when participants finish the survey, we ask them to disclose their familiarity with the movies that they have seen in the survey (``What percentage of movies shown in this survey are familiar to you?''), and we exclude participants who claim more than $40\%$ of the movies they saw were unfamiliar.

 We recruit survey respondents from Amazon Mechanical Turk (MTurk). We send out the survey to $400$ participants. After applying the quality-checking mechanisms, we exclude $16$ participants from our results. The demographics of our survey participants are as follows: $61.2\%$ indicated male and $38.8\%$ indicated female (no participant preferred not to disclose gender); the majority of participants' age fall into the ranges of 21 to 30 ($56.3\%$) or 31 to 40 ($30.7\%$), with $6.0\%$ fall between 41 and 50, and the rest is older than 50.

\section{User Study: Results}
We now analyze the effectiveness of our evaluation score using the collected human ratings, and answer the three research questions in Section~\ref{sec:human}.

\subsection{Is the proposed score correlated with real human judgment? (RQ1)}

\setlength\extrarowheight{6pt}
\begin{table}[!t]
\centering
\resizebox{\columnwidth}{!}{
\begin{tabular}{|ll|l|l|l|l|l|l|l|}
\hline
\multicolumn{2}{|l|}{} & Explainability & Informativeness & Effectiveness & Persuasiveness & Transparency & Trustworthiness & Satisfaction \\ \hline
\multicolumn{1}{|l|}{\multirow{2}{*}{\rotatebox[origin=c]{90}{\footnotesize \textbf{Proposed}}}} & $\cfsym$ & \textbf{0.4436} & \textbf{0.4633} & \textbf{0.3653} & 0.3414 & 0.3701 & \textbf{0.3664} & \textbf{0.3499} \\ \cline{2-9} 
\multicolumn{1}{|l|}{} & $\acfsym$ & 0.4194 & 0.4199 & 0.3404 & \textbf{0.3467} & 0.3347 & 0.3424 & 0.3408 \\ \hline
\multicolumn{1}{|l|}{\multirow{2}{*}{\rotatebox[origin=c]{90}{\footnotesize \textbf{Baseline}}}} & \itemsim & 0.3128 & 0.2619 & 0.2224 & 0.2394 & 0.2127 & 0.3048 & 0.2304 \\ \cline{2-9} 
\multicolumn{1}{|l|}{} & \genresim & 0.3553 & 0.2943 & 0.3221 & 0.2197 & \textbf{0.4554} & 0.3295 & 0.2957 \\ \hline
\end{tabular}
}
\caption{Pairwise correlation between each type of evaluation score and the real human ratings. The higher is better.}
\label{tab:corr}
\end{table}

We compute the correlation between each evaluation score and the average human rating (over participants) across all questions. For both evaluation score and human rating, the larger is the better the explanation quality is. Therefore a more positive correlation means the evaluation score can represent real human ratings better. Table~\ref{tab:corr} shows the pairwise correlation between each evaluation score and the human rating from each assessing dimension. We have two main observations. \textit{First}, $\cfsym$ and $\acfsym$ have a higher correlation than the other two baselines in 6 out of 7 assessing dimensions. We will explain why our method is underperformed by ~\genresim{} in the \textit{Transparency} dimension shortly after. \textit{Second}, $\cfsym$ and $\acfsym$ share a similar correlation, which suggests that the approximation in $\acfsym$ does not significantly downgrade the effectiveness in serving as a proxy of human judgments. ($\acfsym$ even slightly outperforms $\cfsym$ in \textit{Persuasiveness}. However the difference is small.) Hence, in practice $\acfsym$ can be used as a reasonably good evaluation score while saving much computational cost.

\para{Understand Transparency.} To understand why our proposed scores are underperformed in the \textit{Transparency} dimension, we investigate the explanations that users wrote when rating \textit{Transparency}. We find that, when giving a high rating on \textit{Transparency}, many users regarded an explanation's easiness to be understood as the deciding factor. For example, ``It's easy to understand the system.'' (P302), ``I can nearly understand how the system works.''(P174), ``Easy to understand.'' (P143) \etc{} And many users mentioned that sharing similar genres (between the recommended movie and the explaining movies) eased their understanding. For example, ``They almost match the movies before for genre. I got it.'' (P81), ``The explanations share the similar genre. It makes sense to me.'' (P324), ``The genres are matched. I can understand it.'' (P253) \etc{} Hence it is possible that \textit{Transparency} is a dimension that users are naturally inclined to associate with genre similarity because of the intuitiveness.

\setlength\extrarowheight{6pt}
\begin{table}[!t]
\centering
\resizebox{\columnwidth}{!}{
\begin{tabular}{|ll|l|l|l|l|l|l|l|}
\hline
\multicolumn{2}{|l|}{} & Explainability & Informativeness & Effectiveness & Persuasiveness & Transparency & Trustworthiness & Satisfaction \\ \hline
\multicolumn{1}{|l|}{\multirow{2}{*}{\rotatebox[origin=c]{90}{\footnotesize \textbf{Proposed}}}} & $\cfsym$ & \textbf{0.0093} & \textbf{0.0101} & \textbf{0.0089} & \textbf{0.0059} & 0.0032 & \textbf{0.0018} & \textbf{0.0011} \\ \cline{2-9} 
\multicolumn{1}{|l|}{} & $\acfsym$ & 0.0096 & 0.0113 & 0.0093 & 0.0061 & 0.0036 & 0.0022 & 0.0013 \\ \hline
\multicolumn{1}{|l|}{\multirow{2}{*}{\rotatebox[origin=c]{90}{\footnotesize \textbf{Baseline}}}} & \itemsim & 0.0105 & 0.0142 & 0.0114 & 0.0067 & 0.0039 & 0.0059  & 0.0015 \\ \cline{2-9} 
\multicolumn{1}{|l|}{} & \genresim & 0.0141 & 0.0121 & 0.0139 & 0.0081 & \textbf{0.0029} & 0.0031 & 0.0037 \\ \hline
\end{tabular}
}
\caption{Test mean squared error (MSE) of linear regression models that predict human ratings from evaluation scores. The lower MSE is, the better the evaluation score can serve as a predictor of real human ratings.}
\label{tab:mse}
\end{table}

\setlength\extrarowheight{0pt}
\begin{table}[!t]
\centering
\resizebox{\columnwidth}{!}{
\begin{tabular}{|l|l|l|l|l|l|l|l|}
\hline
 & Explainability & Informativeness & Effectiveness & Persuasiveness & Transparency & Trustworthiness & Satisfaction \\ \hline
\begin{tabular}[c]{@{}l@{}}High-$\cfsym$ \\ Mean $\pm$ Std\end{tabular} & \begin{tabular}[c]{@{}l@{}}4.119 $\pm$ 0.953 \end{tabular} & \begin{tabular}[c]{@{}l@{}}3.995 $\pm$ 1.126\end{tabular} & \begin{tabular}[c]{@{}l@{}}4.115 $\pm$ 0.975\end{tabular} & \begin{tabular}[c]{@{}l@{}}4.221 $\pm$ 0.878\end{tabular} & \begin{tabular}[c]{@{}l@{}}4.341 $\pm$ 0.776\end{tabular} & \begin{tabular}[c]{@{}l@{}}4.290 $\pm$ 0.811\end{tabular} & \begin{tabular}[c]{@{}l@{}}4.331 $\pm$ 0.750\end{tabular} \\ \hline
\begin{tabular}[c]{@{}l@{}}Low-$\cfsym$ \\ Mean $\pm$ Std\end{tabular} & \begin{tabular}[c]{@{}l@{}}3.571 $\pm$ 1.101\end{tabular} & \begin{tabular}[c]{@{}l@{}}3.594 $\pm$ 1.140\end{tabular} & \begin{tabular}[c]{@{}l@{}}3.543 $\pm$ 1.064\end{tabular} & \begin{tabular}[c]{@{}l@{}}3.737 $\pm$ 1.107\end{tabular} & \begin{tabular}[c]{@{}l@{}}3.756 $\pm$ 1.078\end{tabular} & \begin{tabular}[c]{@{}l@{}}3.664 $\pm$ 1.091\end{tabular} & \begin{tabular}[c]{@{}l@{}}3.824 $\pm$ 1.080\end{tabular} \\ \hline
t-statistic & 6.518 & 4.297 & 6.422 & 2.0424 & 5.928 & 7.529 & 5.897 \\ \hline
p-value & 2.470e-10 & 1.311e-5 & 4.223e-10 & 5.978e-9 & 3.992e-12 & 6.851e-13 & 7.027e-9 \\ \hline
\end{tabular}
}
\caption{One-tailed (upper tail) paired t-statistics on user ratings on high- and low-$\cfsym$ explanations. The high-$\cfsym$ explanations are rated higher by users than low-$\cfsym$ explanations with statistical significance ($p < 0.05$) on all assessing dimensions.}
\label{tab:test}
\end{table}

\subsection{Can the proposed score be used to predict human judgment? (RQ2)}
In addition to closely correlating with human ratings, an ideal evaluation score should also serve as a good predictor of human ratings. To this end, we build linear regression models that predict the human ratings from the evaluation score. We randomly split the collected survey data into $70\%$ training set and $30\%$ test set, train linear regression models on the training set, and report the test mean squared error (MSE) on the test set in Table~\ref{tab:mse}. The lower test MSE is, the better the evaluation score can serve as a predictor of real human ratings. The observations are mainly consistent with the previous correlation results: $\cfsym$ and $\acfsym$ share a similar range of test MSE, which is lower than the other two baselines except in the \textit{Transparency} dimension where \genresim{} performs slightly better. The result shows that both $\cfsym$ and $\acfsym$ can be used as reasonably good predictors of real human ratings.

\subsection{Is an explanation with a high proposed score considered good by humans? (RQ3)}
Another way to show the evaluative power of $\cfsym$ is to compare the human ratings on explanations with high and low $\cfsym$ (\ie the explanation candidate with the highest and the lowest $\cfsym$ among all candidates, as mentioned in Section~\ref{subsec:generate}). To this end, we perform paired t-test on the human ratings assigned to high- and low-$\cfsym$ explanations. We use the one-tailed (upper tail) setting where the null hypothesis is the high-$\cfsym$ and low-$\cfsym$ explanations share the same average rating, and the alternative hypothesis is the average human rating on high-$\cfsym$ explanations is greater than low-$\cfsym$. Table~\ref{tab:test} shows the corresponding t-statistics. Across all assessing dimensions, the resulting p-value is less than 0.05, indicating we can reject the null hypothesis. Therefore the high-$\cfsym$ explanations are rated higher by users than low-$\cfsym$ explanations with statistical significance. In addition, similar results are found between high- and low-$\acfsym$ explanations, thus we omit the results for brevity.

Similarly, we also compare explanations assessed as high by $\cfsym$ and by the other two baselines. We again perform a similar paired t-test. Table~\ref{tab:compare} shows the results that compare high-$\cfsym$ explanations to high-\itemsim{} and high-\genresim{} explanations. For both baselines, we find that high-$\cfsym$ explanations are rated higher by users than the other two baselines with statistical significance ($p < 0.05$) on all assessing dimensions.

\setlength\extrarowheight{0pt}
\begin{table}[!t]
    \begin{subtable}[h]{\textwidth}
        \centering
        \resizebox{\columnwidth}{!}{
         \begin{tabular}{|l|l|l|l|l|l|l|l|}
        \hline
         & Explainability & Informativeness & Effectiveness & Persuasiveness & Transparency & Trustworthiness & Satisfaction \\ \hline
        \begin{tabular}[c]{@{}l@{}} High-$\cfsym$ \\ Mean $\pm$ Std\end{tabular} & \begin{tabular}[c]{@{}l@{}}4.119 $\pm$ 0.953 \end{tabular} & \begin{tabular}[c]{@{}l@{}}3.995 $\pm$ 1.126\end{tabular} & \begin{tabular}[c]{@{}l@{}}4.115 $\pm$ 0.975\end{tabular} & \begin{tabular}[c]{@{}l@{}}4.221 $\pm$ 0.878\end{tabular} & \begin{tabular}[c]{@{}l@{}}4.341 $\pm$ 0.776\end{tabular} & \begin{tabular}[c]{@{}l@{}}4.290 $\pm$ 0.811\end{tabular} & \begin{tabular}[c]{@{}l@{}}4.331 $\pm$ 0.750\end{tabular} \\ \hline
        \begin{tabular}[c]{@{}l@{}} High-\itemsim \\ Mean $\pm$ Std\end{tabular} & \begin{tabular}[c]{@{}l@{}}3.862 $\pm$ 0.774\end{tabular} & \begin{tabular}[c]{@{}l@{}}3.847 $\pm$ 0.820\end{tabular} & \begin{tabular}[c]{@{}l@{}}3.857 $\pm$ 0.866\end{tabular} & \begin{tabular}[c]{@{}l@{}}3.972 $\pm$ 0.785\end{tabular} & \begin{tabular}[c]{@{}l@{}}3.981 $\pm$ 0.762\end{tabular} & \begin{tabular}[c]{@{}l@{}}3.889 $\pm$ 0.818\end{tabular} & \begin{tabular}[c]{@{}l@{}}3.949 $\pm$ 0.793\end{tabular} \\ \hline
        t-statistic & 3.975 & 1.877 & 3.455 & 4.077 & 5.646 & 6.229 & 6.219 \\ \hline
        p-value & 4.796e-5 & 0.031 & 4.223e-10 & 3.315e-4 & 2.564e-8 & 1.204e-9 & 1.272e-9 \\ \hline
        \end{tabular}
        }
       \caption{Between high-$\cfsym$ and high-\itemsim{} explanations.}
       \label{tab:compare1}
    \end{subtable}
    \\
    \begin{subtable}[h]{\textwidth}
        \centering
        \resizebox{\columnwidth}{!}{
        \begin{tabular}{|l|l|l|l|l|l|l|l|}
        \hline
         & Explainability & Informativeness & Effectiveness & Persuasiveness & Transparency & Trustworthiness & Satisfaction \\ \hline
        \begin{tabular}[c]{@{}l@{}} High-$\cfsym$ \\ Mean $\pm$ Std\end{tabular} & \begin{tabular}[c]{@{}l@{}}4.119 $\pm$ 0.953 \end{tabular} & \begin{tabular}[c]{@{}l@{}}3.995 $\pm$ 1.126\end{tabular} & \begin{tabular}[c]{@{}l@{}}4.115 $\pm$ 0.975\end{tabular} & \begin{tabular}[c]{@{}l@{}}4.221 $\pm$ 0.878\end{tabular} & \begin{tabular}[c]{@{}l@{}}4.341 $\pm$ 0.776\end{tabular} & \begin{tabular}[c]{@{}l@{}}4.290 $\pm$ 0.811\end{tabular} & \begin{tabular}[c]{@{}l@{}}4.331 $\pm$ 0.750\end{tabular} \\ \hline
        \begin{tabular}[c]{@{}l@{}} High-\genresim \\ Mean $\pm$ Std\end{tabular} & \begin{tabular}[c]{@{}l@{}}3.764 $\pm$ 1.018\end{tabular} & \begin{tabular}[c]{@{}l@{}}3.677 $\pm$ 1.135\end{tabular} & \begin{tabular}[c]{@{}l@{}}3.705 $\pm$ 1.122\end{tabular} & \begin{tabular}[c]{@{}l@{}}4.014 $\pm$ 0.887\end{tabular} & \begin{tabular}[c]{@{}l@{}}3.963 $\pm$ 0.925\end{tabular} & \begin{tabular}[c]{@{}l@{}}3.995 $\pm$ 0.893\end{tabular} & \begin{tabular}[c]{@{}l@{}}4.013 $\pm$ 0.933\end{tabular} \\ \hline
        t-statistic & 4.594 & 3.664 & 4.812 & 2.775 & 5.184 & 4.200 & 4.415 \\ \hline
        p-value & 3.699e-6 & 1.563e-4 & 1.402e-6 & 2.997e-3 & 2.486e-7 & 1.951e-5 & 7.959e-6 \\ \hline
        \end{tabular}
        }
        \caption{Between high-$\cfsym$ and high-\genresim{} explanations.}
        \label{tab:compare2}
     \end{subtable}
     \caption{Paired t-statistics that compare user ratings on explanations assessed as high by $\cfsym$ and assessed as high by the baselines. High-$\cfsym$ explanations are rated higher by users than the other two baselines with statistical significance ($p < 0.05$) on all assessing dimensions.}
     \label{tab:compare}
\end{table}
\section{Limitations}
\label{sec:limit}
We point out several limitations in our work. \textit{First and foremost}, our method might be considered to be slow because of the need for retraining. Our focus is to explore the direction of using the counterfactual approach to evaluate explanations and to examine how close the resultant metric is related to human judgments. We believe if the direction seems promising, then the future work can improve efficiency, especially as machine unlearning techniques improve. For example, one potential improvement is to replace retraining with influence function~\cite{koh2017understanding}. \textit{Second}, we only consider evaluating explanations generated based on \textit{item-based collaborative filtering} logic mainly because other logic would require different displaying user interfaces and recommendation system design, and therefore it is difficult to compare under a well-controlled setting. \textit{Third}, there might exist some other possible definitions of counterfactual scores other than the one proposed in Eq (\ref{eq:cf}), \eg considering rank change rather than score change. We do not list and examine all of them. \textit{Finally}, we do not build a real recommendation platform, hire participants to actually use it for a certain period, and gather the data from the real user experience.
\section{Related Work}
\label{sec:related}

In addition to the related work already mentioned in Section~\ref{sec:back}, we include more related work in this section.

\para{Counterfactual Psychology.} Psychological research on counterfactual thinking~\cite{mandel2005psychology} starts in the 1970s with studies that compare human memory between counterfactual and factual inference~\cite{carpenter1973extracting,fillenbaum1974information}.
It is popularized by Kahneman and Tversky in the seminal paper~\cite{kahneman1981simulation}, in which they examine and identify the counterfactual thinking in everylife activities, \eg consumer choice, monetary decisions, and career plans, and connected to a wide range of psychological behaviors. Later, researchers show counterfactual thinking might imply causal inference~\cite{epstude2008functional,spellman1999possibility}. Some studies also point out that it serves as a preparative function~\cite{markman2003reflection,roese1994functional} that helps humans plan for the future, which fits the design of using counterfactual thinking in recommender systems to provide actionable advice to users on how to change their recommendations received in the future.

\para{Machine Unlearning.} Our counterfactual approach is closely related to the ongoing research area of machine unlearning, which studies how to quickly obtain the counterfactually trained model. Bourtoule~\etal~\cite{bourtoule2021machine} propose to partition the training dataset into sub-datasets, and train an individual sub-model on each sub-dataset. Then during inference time, the prediction is made by voting from all sub-models. In this framework, removing the impact of a data sample is to discard the corresponding sub-model trained on the sub-dataset that includes that sample. In addition, Golatkar~\etal~\cite{golatkar2020eternal} use information theory based methods to remove information from the trained model weights. Furthermore, Golatkar~\etal~\cite{golatkar2021mixed} assume there is an essential subset in the training dataset that will not be removed. They train the model on both essential and non-essential datasets with separated model weights. Then discarding non-essential model weights becomes relatively fast because they do not need to change the essential model. Researchers also propose to use caching information during training~\cite{wu2020deltagrad}, or design unlearning methods for a particular class of models like tree-based~\cite{brophy2021machine}.

% The goal of machine unlearning is to efficiently remove the impact of training samples on an already trained model without fully retraining it. It concerns exactly how to obtain the counterfactually trained model, which is our main computational bottleneck. Similar to our method, the main challenge of machine unlearning is also how to reduce the computational cost. We believe our method will become more and more efficient as machine unlearning techniques develop. 
% In this work, we focus on examining the effectiveness of counterfactual approach rather than optimizing the computational efficiency.
\section{Conclusion}
\label{sec:conclusion}
We propose an evaluation method using counterfactual logic that quantifies the quality of explanations in a recommender system. Through an extensive user study, we show that the evaluation scores produced by our method are closely correlated with human judgments. It can serve as a reasonably good proxy of human evaluation and replace the slow and expensive human evaluations. In addition, we show that explanations with high proposed scores are rated higher by humans than by the baselines. In summary, the counterfactual approach to evaluating recommendation explanations seems promising.

We hope that our work can bring more attention to the research on evaluations of recommendation explanations. We believe that practitioners and researchers can benefit significantly from effective and practical evaluations in their algorithm developments, as we see in the other areas.

\bibliography{reference}

\begin{thebibliography}{10}

\bibitem{youtubefairness}
Black creators sue youtube, alleging racial discrimination.
\newblock
  \url{https://www.washingtonpost.com/technology/2020/06/18/black-creators-sue-youtube-alleged-race-discrimination},
  2020.

\bibitem{facebookfairness}
Facebook apologizes after a.i. puts ‘primates’ label on video of black men.
\newblock
  \url{https://www.nytimes.com/2021/09/03/technology/facebook-ai-race-primates.html},
  2021.

\bibitem{twitterfairness}
Twitter’s photo-cropping algorithm prefers young, beautiful, and
  light-skinned faces.
\newblock
  \url{https://www.theverge.com/2021/8/10/22617972/twitter-photo-cropping-algorithm-ai-bias-bug-bounty-results},
  2021.

\bibitem{darpa}
Explainable artificial intelligence (xai).
\newblock
  \url{https://www.darpa.mil/program/explainable-artificial-intelligence},
  2017.

\bibitem{aiuk}
Ai in the uk: ready, willing and able?
\newblock
  \url{https://publications.parliament.uk/pa/ld201719/ldselect/ldai/100/100.pdf},
  2018.

\bibitem{chinaprop}
Full translation: China's 'new generation artificial intelligence development
  plan'.
\newblock
  \url{https://www.newamerica.org/cybersecurity-initiative/digichina/blog/full-translation-chinas-new-generation-artificial-intelligence-development-plan-2017},
  2017.

\bibitem{resnick1994grouplens}
Paul Resnick, Neophytos Iacovou, Mitesh Suchak, Peter Bergstrom, and John
  Riedl.
\newblock Grouplens: An open architecture for collaborative filtering of
  netnews.
\newblock In {\em Proc. of CSCW}, 1994.

\bibitem{herlocker2000explaining}
Jonathan~L Herlocker, Joseph~A Konstan, and John Riedl.
\newblock Explaining collaborative filtering recommendations.
\newblock In {\em Proc. of CSCW}, 2000.

\bibitem{sarwar2001item}
Badrul Sarwar, George Karypis, Joseph Konstan, and John Riedl.
\newblock Item-based collaborative filtering recommendation algorithms.
\newblock In {\em Proc. of TheWebConf}, 2001.

\bibitem{zhang2014explicit}
Yongfeng Zhang, Guokun Lai, Min Zhang, Yi~Zhang, Yiqun Liu, and Shaoping Ma.
\newblock Explicit factor models for explainable recommendation based on
  phrase-level sentiment analysis.
\newblock In {\em Proc. of SIGIR}, 2014.

\bibitem{wu2015flame}
Yao Wu and Martin Ester.
\newblock Flame: A probabilistic model combining aspect based opinion mining
  and collaborative filtering.
\newblock In {\em Proc. of WSDM}, 2015.

\bibitem{costa2018automatic}
Felipe Costa, Sixun Ouyang, Peter Dolog, and Aonghus Lawlor.
\newblock Automatic generation of natural language explanations.
\newblock In {\em Proc. of IUI}, 2018.

\bibitem{sharma2013social}
Amit Sharma and Dan Cosley.
\newblock Do social explanations work? studying and modeling the effects of
  social explanations in recommender systems.
\newblock In {\em Proc. of TheWebConf}, 2013.

\bibitem{chaney2015probabilistic}
Allison~JB Chaney, David~M Blei, and Tina Eliassi-Rad.
\newblock A probabilistic model for using social networks in personalized item
  recommendation.
\newblock In {\em Proc. of RecSys}, 2015.

\bibitem{wang2014also}
Beidou Wang, Martin Ester, Jiajun Bu, and Deng Cai.
\newblock Who also likes it? generating the most persuasive social explanations
  in recommender systems.
\newblock In {\em Proc. of AAAI}, 2014.

\bibitem{zhang2015daily}
Yongfeng Zhang, Min Zhang, Yi~Zhang, Guokun Lai, Yiqun Liu, Honghui Zhang, and
  Shaoping Ma.
\newblock Daily-aware personalized recommendation based on feature-level time
  series analysis.
\newblock In {\em Proc. of TheWebConf}, 2015.

\bibitem{chen2016learning}
Xu~Chen, Zheng Qin, Yongfeng Zhang, and Tao Xu.
\newblock Learning to rank features for recommendation over multiple
  categories.
\newblock In {\em Proc. of SIGIR}, 2016.

\bibitem{seo2017interpretable}
Sungyong Seo, Jing Huang, Hao Yang, and Yan Liu.
\newblock Interpretable convolutional neural networks with dual local and
  global attention for review rating prediction.
\newblock In {\em Proc. of RecSys}, 2017.

\bibitem{wu2019context}
Libing Wu, Cong Quan, Chenliang Li, Qian Wang, Bolong Zheng, and Xiangyang Luo.
\newblock A context-aware user-item representation learning for item
  recommendation.
\newblock {\em ACM Transactions on Information Systems}, 37(2):1--29, 2019.

\bibitem{gao2019explainable}
Jingyue Gao, Xiting Wang, Yasha Wang, and Xing Xie.
\newblock Explainable recommendation through attentive multi-view learning.
\newblock In {\em Proc. of AAAI}, 2019.

\bibitem{heckel2017scalable}
Reinhard Heckel, Michail Vlachos, Thomas Parnell, and Celestine D{\"u}nner.
\newblock Scalable and interpretable product recommendations via overlapping
  co-clustering.
\newblock In {\em Proc. of ICDE}, 2017.

\bibitem{he2015trirank}
Xiangnan He, Tao Chen, Min-Yen Kan, and Xiao Chen.
\newblock Trirank: Review-aware explainable recommendation by modeling aspects.
\newblock In {\em Proc. of CIKM}, 2015.

\bibitem{wang2018tem}
Xiang Wang, Xiangnan He, Fuli Feng, Liqiang Nie, and Tat-Seng Chua.
\newblock Tem: Tree-enhanced embedding model for explainable recommendation.
\newblock In {\em Proc. of TheWebConf}, 2018.

\bibitem{chang2016crowd}
Shuo Chang, F~Maxwell Harper, and Loren~Gilbert Terveen.
\newblock Crowd-based personalized natural language explanations for
  recommendations.
\newblock In {\em Proc. of RecSys}, 2016.

\bibitem{wang2018explainable}
Nan Wang, Hongning Wang, Yiling Jia, and Yue Yin.
\newblock Explainable recommendation via multi-task learning in opinionated
  text data.
\newblock In {\em Proc. of SIGIR}, 2018.

\bibitem{kahneman1981simulation}
Daniel Kahneman and Amos Tversky.
\newblock The simulation heuristic.
\newblock Technical report, Stanford Univ CA Dept of Psychology, 1981.

\bibitem{mandel2005psychology}
David~R Mandel, Denis~J Hilton, and Patrizia~Ed Catellani.
\newblock {\em The psychology of counterfactual thinking.}
\newblock Routledge, 2005.

\bibitem{carpenter1973extracting}
Patricia~A Carpenter.
\newblock Extracting information from counterfactual clauses.
\newblock {\em Journal of Verbal Learning and Verbal Behavior}, 12(5):512--521,
  1973.

\bibitem{fillenbaum1974information}
Samuel Fillenbaum.
\newblock Information amplified: Memory for counterfactual conditionals.
\newblock {\em Journal of Experimental Psychology}, 102(1):44, 1974.

\bibitem{wachter2017counterfactual}
Sandra Wachter, Brent Mittelstadt, and Chris Russell.
\newblock Counterfactual explanations without opening the black box: Automated
  decisions and the gdpr.
\newblock {\em Harv. JL \& Tech.}, 31:841, 2017.

\bibitem{mothilal2020explaining}
Ramaravind~K Mothilal, Amit Sharma, and Chenhao Tan.
\newblock Explaining machine learning classifiers through diverse
  counterfactual explanations.
\newblock In {\em Proc. of FAccT}, 2020.

\bibitem{russell2019efficient}
Chris Russell.
\newblock Efficient search for diverse coherent explanations.
\newblock In {\em Proc. of FAccT}, 2019.

\bibitem{papineni2002bleu}
Kishore Papineni, Salim Roukos, Todd Ward, and Wei-Jing Zhu.
\newblock Bleu: a method for automatic evaluation of machine translation.
\newblock In {\em Proc. of ACL}, 2002.

\bibitem{al2007relationship}
Azzah Al-Maskari, Mark Sanderson, and Paul Clough.
\newblock The relationship between ir effectiveness measures and user
  satisfaction.
\newblock In {\em Proc. of SIGIR}, 2007.

\bibitem{tran2021users}
Thi Ngoc~Trang Tran, Viet~Man Le, M{\"u}sl{\"u}m Atas, Alexander Felfernig,
  Martin Stettinger, and Andrei Popescu.
\newblock Do users appreciate explanations of recommendations? an analysis in
  the movie domain.
\newblock In {\em Proc. of RecSys}, 2021.

\bibitem{zhang2018explainable}
Yongfeng Zhang and Xu~Chen.
\newblock Explainable recommendation: A survey and new perspectives.
\newblock {\em arXiv preprint arXiv:1804.11192}, 2018.

\bibitem{pazzani1999framework}
Michael~J Pazzani.
\newblock A framework for collaborative, content-based and demographic
  filtering.
\newblock {\em Artificial intelligence review}, 13(5):393--408, 1999.

\bibitem{zhao2014we}
Xin~Wayne Zhao, Yanwei Guo, Yulan He, Han Jiang, Yuexin Wu, and Xiaoming Li.
\newblock We know what you want to buy: a demographic-based system for product
  recommendation on microblogs.
\newblock In {\em Proc. of KDD}, 2014.

\bibitem{zhao2016exploring}
Wayne~Xin Zhao, Sui Li, Yulan He, Liwei Wang, Ji-Rong Wen, and Xiaoming Li.
\newblock Exploring demographic information in social media for product
  recommendation.
\newblock {\em Knowledge and Information Systems}, 49(1):61--89, 2016.

\bibitem{vig2009tagsplanations}
Jesse Vig, Shilad Sen, and John Riedl.
\newblock Tagsplanations: explaining recommendations using tags.
\newblock In {\em Proc. of IUI}, 2009.

\bibitem{cramer2008effects}
Henriette Cramer, Vanessa Evers, Satyan Ramlal, Maarten Van~Someren, Lloyd
  Rutledge, Natalia Stash, Lora Aroyo, and Bob Wielinga.
\newblock The effects of transparency on trust in and acceptance of a
  content-based art recommender.
\newblock {\em User Modeling and User-adapted interaction}, 18(5):455, 2008.

\bibitem{pazzani2007content}
Michael~J Pazzani and Daniel Billsus.
\newblock Content-based recommendation systems.
\newblock In {\em The adaptive web}, pages 325--341. Springer, 2007.

\bibitem{hou2019explainable}
Yunfeng Hou, Ning Yang, Yi~Wu, and S~Yu Philip.
\newblock Explainable recommendation with fusion of aspect information.
\newblock {\em World Wide Web}, 22(1):221--240, 2019.

\bibitem{abdollahi2017using}
Behnoush Abdollahi and Olfa Nasraoui.
\newblock Using explainability for constrained matrix factorization.
\newblock In {\em Proc. of RecSys}, 2017.

\bibitem{dhurandhar2018explanations}
Amit Dhurandhar, Pin-Yu Chen, Ronny Luss, Chun-Chen Tu, Paishun Ting,
  Karthikeyan Shanmugam, and Payel Das.
\newblock Explanations based on the missing: Towards contrastive explanations
  with pertinent negatives.
\newblock In {\em Proc. of NeurIPS}, 2018.

\bibitem{goyal2019counterfactual}
Yash Goyal, Ziyan Wu, Jan Ernst, Dhruv Batra, Devi Parikh, and Stefan Lee.
\newblock Counterfactual visual explanations.
\newblock In {\em Proc. of ICML}, 2019.

\bibitem{akula2020cocox}
Arjun Akula, Shuai Wang, and Song-Chun Zhu.
\newblock Cocox: Generating conceptual and counterfactual explanations via
  fault-lines.
\newblock In {\em Proc. of AAAI}, 2020.

\bibitem{wu2021polyjuice}
Tongshuang Wu, Marco~Tulio Ribeiro, Jeffrey Heer, and Daniel~S Weld.
\newblock Polyjuice: Generating counterfactuals for explaining, evaluating, and
  improving models.
\newblock In {\em Proc. of ACL}, 2021.

\bibitem{yang2020generating}
Linyi Yang, Eoin~M Kenny, Tin Lok~James Ng, Yi~Yang, Barry Smyth, and Ruihai
  Dong.
\newblock Generating plausible counterfactual explanations for deep
  transformers in financial text classification.
\newblock In {\em Proc. of COLING}, 2020.

\bibitem{ying2019gnnexplainer}
Rex Ying, Dylan Bourgeois, Jiaxuan You, Marinka Zitnik, and Jure Leskovec.
\newblock Gnnexplainer: Generating explanations for graph neural networks.
\newblock In {\em Proc. of NeurIPS}, 2019.

\bibitem{yeh2019diverse}
Raymond~A Yeh, Alexander~G Schwing, Jonathan Huang, and Kevin Murphy.
\newblock Diverse generation for multi-agent sports games.
\newblock In {\em Proc. of CVPR}, 2019.

\bibitem{verma2020counterfactual}
Sahil Verma, John Dickerson, and Keegan Hines.
\newblock Counterfactual explanations for machine learning: A review.
\newblock {\em arXiv preprint arXiv:2010.10596}, 2020.

\bibitem{ghazimatin2020prince}
Azin Ghazimatin, Oana Balalau, Rishiraj Saha~Roy, and Gerhard Weikum.
\newblock Prince: Provider-side interpretability with counterfactual
  explanations in recommender systems.
\newblock In {\em Proc. of WSDM}, 2020.

\bibitem{tran2021counterfactual}
Khanh~Hiep Tran, Azin Ghazimatin, and Rishiraj~Saha Roy.
\newblock Counterfactual explanations for neural recommenders.
\newblock In {\em Proc. of SIGIR}, 2021.

\bibitem{koh2017understanding}
Pang~Wei Koh and Percy Liang.
\newblock Understanding black-box predictions via influence functions.
\newblock In {\em Proc. of ICML}, 2017.

\bibitem{kaffes2021model}
Vassilis Kaffes, Dimitris Sacharidis, and Giorgos Giannopoulos.
\newblock Model-agnostic counterfactual explanations of recommendations.
\newblock In {\em Proc. of UMAP}, 2021.

\bibitem{cao2015towards}
Yinzhi Cao and Junfeng Yang.
\newblock Towards making systems forget with machine unlearning.
\newblock In {\em Proc. of IEEE S\& P}, 2015.

\bibitem{bourtoule2021machine}
Lucas Bourtoule, Varun Chandrasekaran, Christopher~A Choquette-Choo, Hengrui
  Jia, Adelin Travers, Baiwu Zhang, David Lie, and Nicolas Papernot.
\newblock Machine unlearning.
\newblock In {\em Proc. of IEEE S \& P}, 2021.

\bibitem{gupta2021adaptive}
Varun Gupta, Christopher Jung, Seth Neel, Aaron Roth, Saeed Sharifi-Malvajerdi,
  and Chris Waites.
\newblock Adaptive machine unlearning.
\newblock In {\em Proc. of NeurIPS}, 2021.

\bibitem{guo2019certified}
Chuan Guo, Tom Goldstein, Awni Hannun, and Laurens Van Der~Maaten.
\newblock Certified data removal from machine learning models.
\newblock In {\em Proc. of ICML}, 2020.

\bibitem{als}
Yehuda Koren, Robert Bell, and Chris Volinsky.
\newblock Matrix factorization techniques for recommender systems.
\newblock {\em Computer}, 42(8):30--37, 2009.

\bibitem{sayres2019using}
Rory Sayres, Ankur Taly, Ehsan Rahimy, Katy Blumer, David Coz, Naama Hammel,
  Jonathan Krause, Arunachalam Narayanaswamy, Zahra Rastegar, Derek Wu, et~al.
\newblock Using a deep learning algorithm and integrated gradients explanation
  to assist grading for diabetic retinopathy.
\newblock {\em Ophthalmology}, 126(4):552--564, 2019.

\bibitem{ribeiro2018anchors}
Marco~Tulio Ribeiro, Sameer Singh, and Carlos Guestrin.
\newblock Anchors: High-precision model-agnostic explanations.
\newblock In {\em Proc. of AAAI}, 2018.

\bibitem{ren2017social}
Zhaochun Ren, Shangsong Liang, Piji Li, Shuaiqiang Wang, and Maarten de~Rijke.
\newblock Social collaborative viewpoint regression with explainable
  recommendations.
\newblock In {\em Proc. of WSDM}, 2017.

\bibitem{tintarev2007survey}
Nava Tintarev and Judith Masthoff.
\newblock A survey of explanations in recommender systems.
\newblock In {\em Proc. of ICDE workshop}, 2007.

\bibitem{epstude2008functional}
Kai Epstude and Neal~J Roese.
\newblock The functional theory of counterfactual thinking.
\newblock {\em Personality and social psychology review}, 12(2):168--192, 2008.

\bibitem{spellman1999possibility}
Barbara~A Spellman and David~R Mandel.
\newblock When possibility informs reality: Counterfactual thinking as a cue to
  causality.
\newblock {\em Current Directions in Psychological Science}, 8(4):120--123,
  1999.

\bibitem{markman2003reflection}
Keith~D Markman and Matthew~N McMullen.
\newblock A reflection and evaluation model of comparative thinking.
\newblock {\em Personality and Social Psychology Review}, 7(3):244--267, 2003.

\bibitem{roese1994functional}
Neal~J Roese.
\newblock The functional basis of counterfactual thinking.
\newblock {\em Journal of personality and Social Psychology}, 66(5):805, 1994.

\bibitem{golatkar2020eternal}
Aditya Golatkar, Alessandro Achille, and Stefano Soatto.
\newblock Eternal sunshine of the spotless net: Selective forgetting in deep
  networks.
\newblock In {\em Proc. of CVPR}, 2020.

\bibitem{golatkar2021mixed}
Aditya Golatkar, Alessandro Achille, Avinash Ravichandran, Marzia Polito, and
  Stefano Soatto.
\newblock Mixed-privacy forgetting in deep networks.
\newblock In {\em Proc. of CVPR}, 2021.

\bibitem{wu2020deltagrad}
Yinjun Wu, Edgar Dobriban, and Susan Davidson.
\newblock Deltagrad: Rapid retraining of machine learning models.
\newblock In {\em Proc. of ICML}, 2020.

\bibitem{brophy2021machine}
Jonathan Brophy and Daniel Lowd.
\newblock Machine unlearning for random forests.
\newblock In {\em Proc. of ICML}, 2021.

\end{thebibliography}
\bibliographystyle{unsrt}

% \appendix

\end{document}